\documentclass[sn-mathphys,Numbered,iicol]{sn-jnl}

\usepackage{graphicx}%
\usepackage{multirow}%
\usepackage{amsmath,amssymb,amsfonts}%
\usepackage{amsthm}%
\usepackage{mathrsfs}%
\usepackage[title]{appendix}%
\usepackage{xcolor}%
\usepackage{textcomp}%
\usepackage{manyfoot}%
\usepackage{booktabs}%
\usepackage{algorithm}%
\usepackage{algorithmicx}%
\usepackage{algpseudocode}%
\usepackage{listings}%
\usepackage{orcidlink}

\begin{document}

\title[Article Title]{A NLP Approach to ``Review Bombing'' in Metacritic PC Videogames User Ratings}


\author*[1]{\fnm{Javier} \sur{Coronado-Blázquez}\protect\orcidlink{0000-0001-9643-0125}}\email{j.coronado.blazquez@gmail.com}

\affil*[1]{\orgdiv{Telefónica Tech IoT \& Big Data}, \orgaddress{\street{Ronda de la Comunicación Oeste 1}, \city{Madrid}, \postcode{28050}, \country{Spain}}}


\abstract{Many videogames suffer ``review bombing'' --a large volume of unusually low scores that in may cases do not reflect the real quality of the product-- when rated by users. By taking Metacritic's 50,000+ user score aggregations for PC games in English language, we use a Natural Language Processing (NLP) approach to try to understand the main words and concepts appearing in such cases, reaching a 0.88 accuracy on a validation set when distinguishing between just bad ratings and review bombings. By uncovering and analyzing the patterns driving this phenomenon, these results could be used to further mitigate these situations.}




\maketitle

\section{Introduction}\label{sec1}

Videogames are currently the leading entertainment industry product, exceeding \$193 billion global revenue in 2021, with an annual growth of about 15\% \cite{ey_gaming_next_dimension}. From small indie to multi-million triple-A titles, from first-person shooters to racing simulators and even almost unclassifiable experiences, the current gaming ecosystem is rich and diverse.

As with streaming platforms, the amount of offer can be overwhelming for a casual player. It is only natural that they carry out a little research on the Internet to decide whether a game is worth playing to \cite{Drachen2011OnlyTG}, especially when most of triple-A titles are released at an average price of \$70 \cite{extremetech_ubisoft_ceo}.

Similarly to the Internet Movie Data Base (\href{https://www.imdb.com/}{IMDb}) and \href{https://www.rottentomatoes.com/}{Rotten Tomatoes} for movies and TV shows, or \href{https://www.goodreads.com/}{GoodReads} for books, \href{https://www.metacritic.com/}{Metacritic} was born in 1999 as a score aggregator for different entertainment products, including videogames. For a given title, Metacritic offers an average score --the Metascore-- pondering professional critics from specialized media. Furthermore, any registered user can vote, in a scale of 0 to 10, and leave an optional comment as review \cite{8931866, Johnson2014TheEO}\footnote{Although the Metascore ranges from 0 to 100, in the remaining of the paper we will always refer to the Metascore divided by 10 to equate it to the user score.}.

In recent years, Metacritic has been in the news for the so-called ``review bombing'', this is, titles where the average user score is anomalously low, especially when compared to the Metascore \cite{05c4a57cb23f48a0aa90c52439eddac3}, present a large volume of those, and the user reviews usually lead towards derivative aspects of the game instead of the title quality itself, such as studio marketing campaigns or commercial strategy, change of story focus in sequels, or political and cultural controversies such as feminist perspectives or LGTBIQ+ character representation \cite{dutton2011, dannagal2017}.

Contrary to the expected user reviews, the review bombing tends to be a coordinated action by many individuals in forums such as Reddit or 4chan \cite{Ferguson2020WhoAG}, often with multiple accounts, with the intention of doing as much harm as possible to the game studio, trying to hurt its sales and commercial performance \cite{braithwaite2016, kotaku_steam_review_bombing}.

Indeed, empirical evidence proves that online ratings are not Gaussian-distributed (as naively expected), but rather ``J-shaped'', with a raise at very low scores, indicating that upset users are more prone to leave a review, and this is usually in the first $\mathrm{10^{th}}$ score percentile \cite{Schoenmueller2020ThePO, ASKALIDIS201723, han2020}.

Many authors have utilized automated sentiment analysis to study product reviews and extract insights from the customers' patterns (see, e.g., \cite{singh2022improving, nellutla2021online, mccloskey2024natural, fang2015sentiment}). Additionally, there are several works devoted to the trustworthiness of reviews, and the impact they may have on customers' decisions \cite{FILIERI201646, XIANG2010179, doi:10.1177/1356766718778879, doi:10.1509/jmkr.43.3.345}.

Nevertheless, videogame reviews have only been considered in marketplace reviews (and therefore subject to artificial negative reviews due to problems with the selling platform and/or the delivery process) \cite{agca2023, rodriguez2017purchasing, mudambi2010research, doi:10.1080/0144929X.2018.1456563, chenghsun2012}, but have not been studied in the context of score aggregators.

While Metacritic and other platforms have recognized the problem and started to draw a plan to limit them (from a 48-hour moratorium since the game release date to a weighted user score), deciding whether a user review is ``valid'' or can be labeled as ``review bombing'' is currently a highly-subjective and manual task \cite{forbes_horizon_forbidden_west}. For example, in many cases there are day-one bugs that makes a game almost unplayable in certain platforms, although in most cases the developers fix those problems with free software updates, leading to a user rating increase\footnote{The most notorious case is CD Projekt RED's Cyberpunk 2077, a 2020 highly-anticipated triple-A title, which was released with severe problems to guarantee a stable gaming experience, leading to extremely poor user scores \cite{forbes_cyberpunk_metacritic}. Three years later, subsequent updates have improved the game and, at the time of writing, it holds a 7.1 average user rating.}.

In this paper, we propose a Natural Language Processing (NLP) approach to study this problem, by defining some criteria to create a review-bombing dataset, and analyze the main concepts and words driving the users towards these ratings. This is a novel outlook, as to our knowledge there is no record in literature of such automated, artificial intelligence-based analysis, whereas there are hundreds of articles and web posts discussing this problematic from a sociological and/or economic point of view. In this sense, this work can complement those studies from an analytical perspective, providing a technical framework. This research gap can hopefully ignite an interest on the application of automated artificial intelligence tools to address and properly handle those situations, improving the user experience with unbiased reviews that reflect the actual quality of the product.

The paper is structured as follows: in Section \ref{sec:metacritic_database} we define the criteria used to determine whether a title has suffered review-bombing, and create a dataset of such ratings. Section \ref{sec:models_train} details the training and evaluation process of a NLP algorithm. In Section \ref{sec:wordclouds}, we explore the main concepts and words present in such reviews, as well as further insights. Finally, discussion and conclusions are outlined in Section \ref{sec:discussion}.

\section{Defining a Metacritic ``review bombing'' dataset}
\label{sec:metacritic_database}

Our starting point is a Metacritic PC games review dataset from \href{https://www.kaggle.com/}{Kaggle} \cite{kaggle_metacritic_pc_games_reviews}, comprising a total of 513,250 ratings, for titles spanning from 1995 up to 2023. These include entries for yet-to-be-released games, indicated as 'TBD' either in the date, score o review.

We decide not to translate into English reviews written in other languages, as the translation could not be as precise as desired given the context (not only related to gaming context, which include neologisms, but prone to orthographic errors, or deliberately misleading language)\footnote{This is an inherent limitation of the analysis for some games such as Blizzard's multi-player shooter Overwatch 2, where two thirds of the reviews are written in Chinese --with 97\% of them negative-- due to the widespread shutdown of Blizzard games caused by the ending of NetEase-Blizzard agreement in China, causing players to lose access to their accounts and the ability to play on a national server \cite{ign_overwatch_2_review}.}. English reviews are detected with the \texttt{langdetect} library \cite{ercdidip2022}.

Additionally, being them our main interest, only user reviews are taken into account. Although professional critics are also included in the data, typically a volume of x10--300 times more user ratings are found per title when compared to the specialized media.

Metacritic establishes a minimum 4 ratings to assign a score (either user score or Metascore). Lots of little, modest games are usually not reaching this threshold for users, and therefore will not be taken into account in our sample. Likewise, lots of indie titles lack a minimum of 4 professional reviews. Nevertheless, review bombing is known to happen mainly for triple-A games developed by big studios, which are well known in the community and have a good volume of both professional and user reviews.

We define review bombing candidates when a title meets the following requirements:

\begin{enumerate}
    \item has both a Metascore and user score reported;

    \item has a minimum of 5 user reviews in English;

    \item has a discrepancy between the Metascore and average user rating larger or equal than 4.0 points, in favour of the professional critics.

\end{enumerate}

For example, if the Metascore for a title meeting the two first requirements is 7.0, we will mark it as potentially review-bombing candidate if the average user score is below 3.0. This points towards a very large discrepancy between professional and general reviews.

\begin{figure}
\centering
\includegraphics[width=\linewidth]{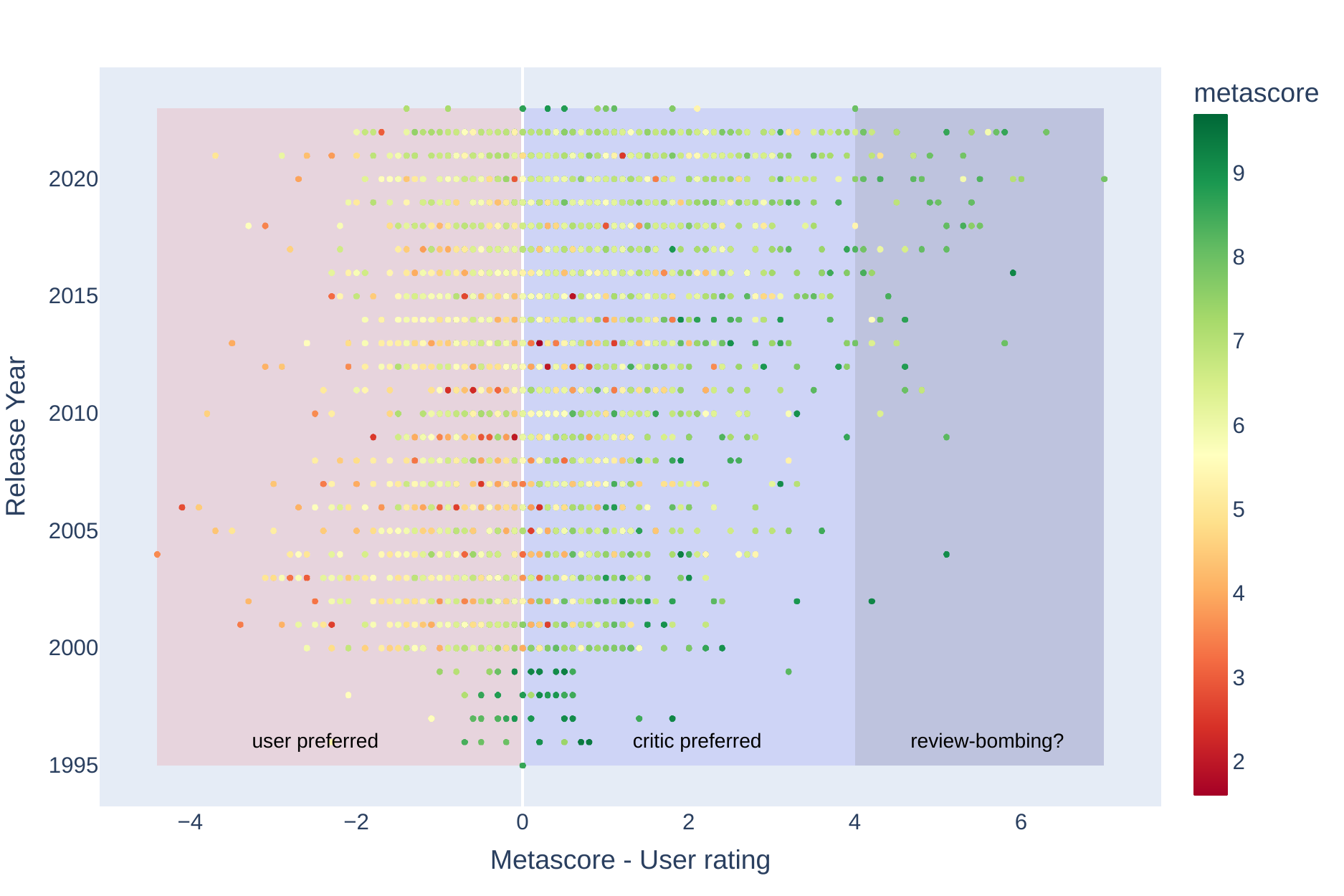}
\caption{Metascore -- user rating scores for PC games (1995--2023). Shaded areas show user-preferred games (average user score $>$ Metascore), critic-preferred titles (average user score $<$ Metascore), and potential games which have suffered review bombing (Metascore minus average user score $>$ 4.0). Colour code traces the Metascore.}
\label{fig:plotly_metacritic}
\end{figure}

Despite the 4.0 point threshold being an \textit{ad hoc} choice, we believe it also possesses two additional advantages; firstly, it removes critics' low-scored games -- if we chose a 3.0 as threshold, a game with a Metascore of 4.0 and average user score of 1.0 would meet the criteria, although this case would not be as clearly review bombing as the 7.0 vs. 3.0 example we presented before. In second place, as presented in Figure \ref{fig:plotly_metacritic}, there are no games exceeding 4 points of difference between average Metascores and user scores in favour of the latter (pointing to ``cult titles''). The fact that this large gap exists only in the opposite case is a symptom of review bombing.

From Figure \ref{fig:plotly_metacritic} some conclusions can already be extracted. When analyzing the release year, we observe how review bombing --as defined with our conservative criteria-- was almost non-existing before $\sim$2010, while it dramatically increases from 2015--2016 onwards. As we pointed before, no title exceeds a 4-point difference in the case of unfavourable professional critics, while in the case of review bombing candidates many titles with more than 5 points of difference are found, with a maximum of a 7-point discrepancy\footnote{The record is held by Tom Clancy's The Division 2: Warlords of New York, with a Metascore of 7.9 and an average user score of 0.9 at the time of writing.}.

With these criteria, our final sample of potential games which have suffered review bombing consists of 52,863 individual reviews for a total of 68 PC games --an average 777 reviews per game, while in the case of non-potential review bombing titles this value drops to 80, across 5,152 titles--. As we mentioned before, review bombing tends to be a coordinated action across individual users, which substantially increases the average number of reviews per game.

In Table \ref{tab:examples_bombing} we show examples of review bombing and examples of just negative reviews, for user scores below 3 points.

\begin{table}[!ht]
\centering
\begin{tabular}{{ |p{0.46\linewidth}|p{0.46\linewidth}| }}
\hline
\textbf{Review bombing} & \textbf{No Review bombing} \\
\hline
\hline
\textcolor{black}{I'm just here to help that score get lower huehuehuehuehuehuehuehhuehuehuehue}
 & \textcolor{black}{literally unplayable. Enemy console players with cheating aim assist will always be better than you. You cannot compete with the trash aim assist system that pull the bullets to the head} \\
\hline
\textcolor{black}{Boycott Blizzard for squirting out this garbage. Until everyone wakes up and STOPS giving this company money, the games they make will NEVER get any better. Warcraft 3 Reforged is proof of this.}
 & \textcolor{black}{It's fun game at the first, but it's so boring when you reach high level. Plain story spesific at 2.1 version, bad character kit design, lack of content and anniversary reward is suck. This devloper become so greddy at this time. I not recomend this game} \\
\hline
\textcolor{black}{Thanks for ruining my childhood you poop faces [x6]} & \textcolor{black}{Awfuly boring gameplay, primitive graphics, many connections errors...} \\
\hline
\textcolor{black}{I'm just scoring low so hopefully Blizzard will wake up and realize that no one likes DRM, and no one wants 10 expansions until they make the next one...Buy to win ******** lame man...Just lame.} & \textcolor{black}{``We continue to investigate reports of ongoing connection issues. Thank you for your patience!'' The gaming experience of this game is their server crashes daily and all my characters progress got rolled back} \\
\hline
\end{tabular}
\caption{\textcolor{black}{Examples of negative Metacritic user reviews, for review bombing (left column) and no review bombing (right column). See text for more details.}}
\label{tab:examples_bombing}
\end{table}

With these reviews, we perform a tokenization of the texts with \texttt{spacy} \cite{ines_montani_2023_10009823}, and create a vector matrix with a \texttt{TfidfVectorizer} from \texttt{scikit-learn} \cite{scikit-learn} using 1000 max features\footnote{We also tried using 500 and 2000 features and the results were very similar.}. With this matrix, we will be able to train different NLP models to address the importance of different words and concepts in these potential review bombings.

\section{NLP models training and evaluation}
\label{sec:models_train}
Once our dataset of potential review bombings is built, we perform a machine-learning (ML) approach with NLP algorithms, aiming to understand the importance of words (lemmas) in those reviews, and to distinguish them from simply bad reviews with no harming intention.

\begin{figure}
\centering
\includegraphics[width=0.49\linewidth]{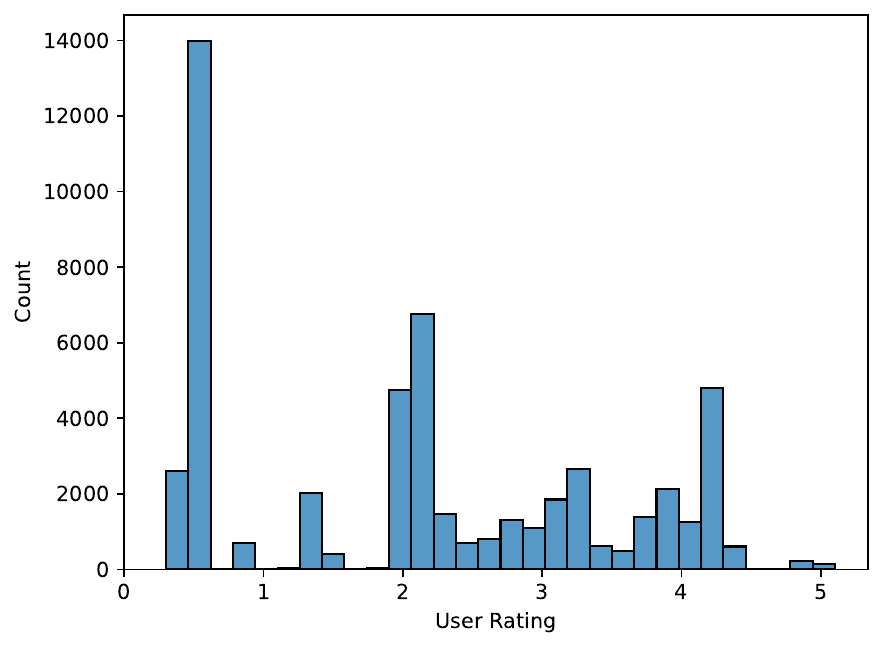}
\includegraphics[width=0.49\linewidth]{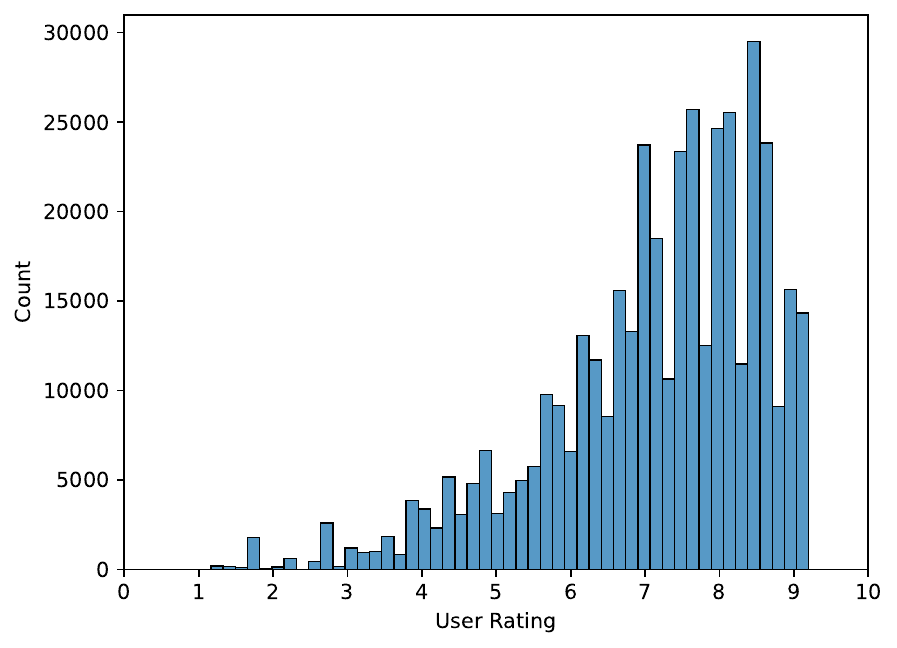}
\caption{User ratings for Metacritic PC game reviews in the sample of potential review bombing titles (left panel) and the sample of non-potential review bombing titles (right panel). See text for details.}
\label{fig:distribution_scores}
\end{figure}

We define two class labels, depending on whether an individual user review is equal or less than 1.0. This seems an excessively aggressive approach, as reviews with a score of 2.0 may be review bombing. Nevertheless, when taking a look at the user score distribution for our review bombing sample, as shown in Figure \ref{fig:distribution_scores}, there is a clear peak below 1.0, as most of the review bombing users aim to assign scores as low as possible. In the same figure, we show the user score distribution for non-bombing reviews, where just $\sim$1.7\% of scores are below 3.0.

The resulting data is divided in a train/test split with 80/20 proportions and stratification on the label frequency. Four models are considered for the classification task: Logistic Regression (LR), Random Forest (RF), Gradient Boosting (GB) and Multinomial Naive Bayes (MNB), all loaded from \texttt{scikit-learn} \cite{scikit-learn}. To ensure the maximum performance of each of them while guaranteeing an unbiased score estimation, a grid search hyperparameter optimization with 5-fold cross-validation is executed.

\begin{figure}[!ht]
\centering
\includegraphics[width=0.75\linewidth]{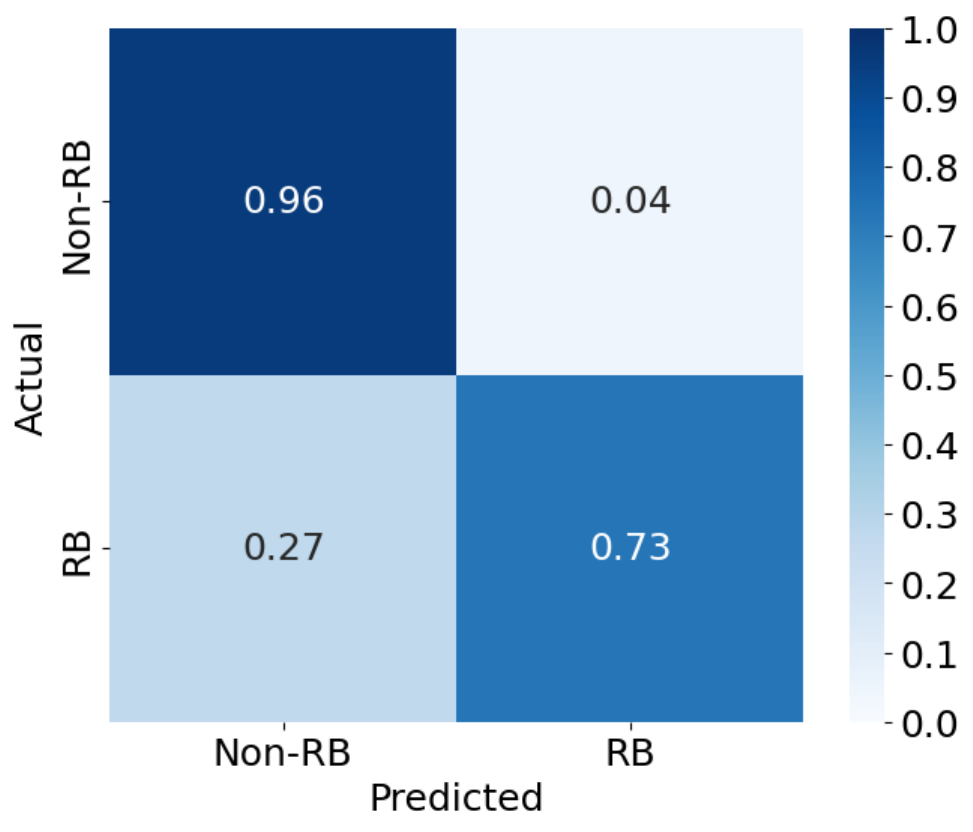}
\caption{Normalized confusion matrix of the MNB best NLP model, for Review Bombing (RB) and Non-Review Bombing (Non-RB).}
\label{fig:confusion_matrix}
\end{figure}

The best model is the MNB ($\alpha_{opt}=0.01$), with a 0.88 accuracy on the validation set. Recall, precision and \textcolor{black}{F1 macro averages are 0.84, 0.88 and 0.86 respectively}. The rest of models present accuracies between 0.81 and 0.85, indicating that they are all competitive for this dataset. The confusion matrix for the MNB model is shown in Figure \ref{fig:confusion_matrix}.

There are almost 7x more false negatives than false positives, i.e., the model assigns 27\% of actual review bombing ratings to the ``normal'' class, yet only 4\% of the opposite. This suggests that review bombing language is more complex than standard reviews, and that, while some user reviews are assigned scores below 1.0 in critic's acclaimed titles, review bombing only exceptionally assigns higher scores.

\section{Insights on review bombing main drivers}
\label{sec:wordclouds}

With a trained ML algorithm with satisfactory metrics on the validation set, we can rank the importance of words and concepts appearing in those ratings flagged as review bombing.

\begin{figure}
\centering
\includegraphics[width=\linewidth]{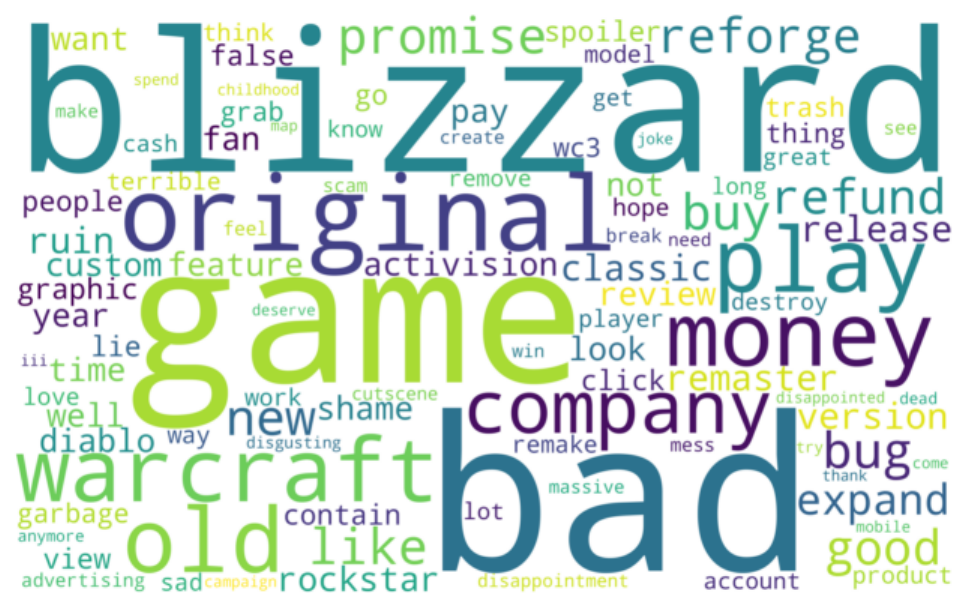}
\caption{Wordcloud for the most relevant concepts in review bombing user ratings. See text for more details.}
\label{fig:wordcloud}
\end{figure}

To do so, we extract the conditional probability from the vector matrix for each of the feature words. By sorting these, a ranking of the most relevant concepts is obtained. In order to enhance the visualization of concepts for the human eye, we present those results as a wordcloud in Figure \ref{fig:wordcloud}, where the size of each word is proportional to its importance.

As seen in the figure, most non-obvious words (ignoring ``game'', ''bad'', etc.) point towards different groups of concepts:

\begin{enumerate}
    \item \textbf{Individual companies and studios:} words such as ``blizzard'',``activision'' or ``rockstar'' indicate that many users target titles developed by these companies. Indeed, many games in the sample are developed by these studios, such as Diablo III, World of Warcraft (and expansions), Grand Theft Auto Remastered or Tom Clancy's The Division 2. Indeed, one of the most important features is ``company'', pointing towards dislike of these individual studios.

    \item \textbf{Originality and content:} concepts like ``original'', ``classic'', ``new'', ``old'', ``remaster'' or ``childhood'' point out to upsetting of users' expectations with the originality of newly produced games, which in many cases are remasters or remakes of mid-90s or early 2000s titles with widespread critical acclaim. New approaches in artistic or gameplay direction, updated scripts or bad ports are usually vectors of review bombing. Many of these users may experiment ``gaming nostalgia'' \cite{MaC1317}, having played the original titles when originally released, and finding the new title unable to meet the expectations of their childhood memories.

    \item \textbf{Economic aspects:} words such as ``money'', ``refund'', ``spend'' or ``cash'' suggests financial motives driving review bombing. This is especially the case for sequels and paid downloadable content (DLC) expanding the base game, where users consider the additions are not worth the retail price. On the other hand, micro-transactions and so-called ``pay-for-win'' are a common monetization strategy --especially in online multiplayer titles-- which is usually not well received by users \cite{GIBSON2022107219}. 

    \item \textbf{Sentiment analysis:} vocabulary including ``garbage'', ``terrible'', ``disgusting'', ``trash'' points towards very negative, extreme feelings towards the product. Within the field of evaluation theory, this is known as ``value-based'' ratings, depending on individual opinions which are influenced by sociocultural factors, in opposition to the fact-based ones which are founded on verifiable facts \cite{datta2007}. This indicates that most of these reviews do not criticize objective aspects, but are rather based upon personal feelings towards the game content.

    \item \textbf{Frustration:} many of the review bombings present words like ``false'', ``promise'', ``shame'', ``hope'', ``lie'', etc. Complementary to the sentiment analysis, this suggests many users perceive some titles as a ``betrayal'' to their expectations. The NLP model finds that reviews containing these words are those most prone to be review bombing, as user groups are genuinely deceived and hope that flooding the scores with near-zero ratings the industry will try to avoid similar situations in the future to prevent sales drops.
\end{enumerate}

It is very interesting to note no political or LGTBIQ+ keywords are found as drivers in this PC dataset. This is actually expected when taking a look to the games and genres in which review bombing is found in PC, being most of them first-person shooters (FPS), massively multiplayer online role-playing game (MMORPG) or action titles, with no such content. Although some recent titles such as Dead Space Remake or Resident Evil 4 Remake have been subject to some review bombing for this reason \cite{gamingbible_dead_space_remake_review_bomb, thegamer_resident_evil_4_remake_review_bombing}, it has been a very little share of ratings, which keeps the user score and Metascore discrepancy below 1 point, therefore not having any relevance at all in the context of score-driven sales trends.

Games with relevant review bombing for political and/or LGTBIQ+ content most notably include Horizon Forbidden West: Burning Shores and The Last of Us Part II \cite{cantone2023ideologydriven}, Sony Playstation exclusives. In such cases, an overwhelming majority of the low-score user reviews were centered on so-called ``woke'' content \cite{Cammaerts2022}, such as feminine main characters, gender-inclusive language, and multi-racial, minorities and/or LGTBIQ+ representation. Multiple zero- or one-point reviews of such titles were just the phrase ``go woke, go broke'', usually employed as slogan calling for boycott of entertainment products featuring such content \cite{rollingstone_woke_companies_profit}. Future work foresees extending this analysis to console-exclusive titles to understand the impact of these reviews in the NLP model.

\section{Discussion and conclusions}
\label{sec:discussion}

In this work, we have studied online score aggregator Metacritic reviews to study the so-called ``review bombing'' phenomenon, in which a large group of users assign unusally low scores to a product with critical acclaim. This has been an increasing concern in the videogame industry, with multiple episodes in the last years.

By using a Kaggle Metacritic dataset comprising a total of 513250 individual ratings for PC games, we limited to user scores written in English, and defined criteria to limit the results to those which potentially have suffered review bombing, establishing a threshold of at least 4.0 point difference between the Metascore and average user score. The final dataset consisted on 52,863 reviews for 68 PC titles.

After cleaning and tokenizing the texts, we trained a Natural Language Processing (NLP) machine learning model, in order to understand the main words and concepts driving review bombing user comments. We trained four different algorithms, namely Logistic Regression, Random Forest, Gradient Boosting and Multinomial Naive Bayes. The latter turned out to be the better performing when trained with a train/test split of 80/20\% with a grid search hyperparameter optimization and 5-fold cross-validation, reaching a global accuracy of 0.88.

By ranking the conditional probabilities of the tokenized text, we were able to gain insights on the main drivers of review bombing. We found 5 groups of concepts: individual companies, originality and content, economic aspects, sentiment analysis and frustration. By taking into account both these families and the individual words the model found to be most relevant, online aggregators can take actions against such situations.

What actions are those is a matter of discussion. Suppression as a definitive strategy to fight review bombing can be a double-edged sword: if a review bombing phenomenon has transcended the gaming community and reached widespread media, there has already been damage to the platform's reputation and prestige. In that situation, an immediate and drastic intervention can be perceived as unwarranted meddling and a signal that the evaluative tool lacks impartiality in its assessment of the subject \cite{graves2017}.

Additionally, understanding the underlying drivers of these reviews --especially when focusing on technical issues or lack of and/or repetitive content-- can be used by studios to improve user experience and as feedback for developers \cite{nilanjan2023}. Indeed, studies of review bombs can be used as an opportunity for further theoretical advancements in the field of information management \cite{Venkatesh2013BridgingTQ}. While it is a common occurrence in social media, the concept of meta-opinion is a relatively novel concept in the realm of online reviews, deviating from the standard model where users provide reviews solely about the product itself. Instead, meta-opinions offer a distinctive perspective by connecting and building upon prior, value-based opinions related to the product. Their primary function is to either bolster or challenge previous messages, thereby contributing to the evaluation of the overall credibility of product reviews \cite{Schwandt2007JudgingIB}.

The analysis we have conducted can be very easily extended to other game score aggregators like \href{https://store.steampowered.com/}{Steam} and even others product user reviews platforms such as IMDb or Goodreads. This way, not only can content moderators benefit from the understanding of underlying drivers, but a transversal analysis can undercover further sociocultural and demographic patterns depending on the platform, product type, etc.

This is especially relevant for political and/or LGTBIQ+ content-based review bombing (``go woke, go broke''), which are not present in PC games due to the most notorious cases happening in console-exclusive titles such as Horizon Forbidden West: Burning Shores or The Last of Us Part II. The impact of such content can significantly alter the conclusions of this analysis, most likely adding a sixth category on the insights groups of concepts.

In this sense, a further consideration for future extensions of this work is the influence of the review date in relation to the game's release date. As an example of such effect, at the time of writing The Last of Us Part II holds a 9.3 Metascore and 5.8 user score. Yet, during the first days after release, the gap widened to a 9.5 Metascore and 3.4 user score \cite{forbes_last_of_us_part_2}. This suggests very early, negative reviews are a coordinated action targeted towards damaging the game reputation, while players with a more fact-based approach tend to spend real time playing the title before delivering a verdict \cite{cantone2023ideologydriven}. Such analysis presents a problem with Metacritic criteria changes precisely aiming to combat review bombing, in which many of these reviews are now removed and a minimum 48h period is left before accepting user reviews.

Additionally, the demographic of the user, namely sex and age, can be very relevant. In the case of Metacritic this information is not reported, yet in other platforms suchs as IMDb it is. As an example of such importance, HBO's The Last of Us --adaptation of the title of the same name-- received review bombing in its third episode for a LGTBIQ+ plot not explicitly showed (yet implicit) in the original game. When looking at the score by demographics, there is a big difference between the ratings by male or female users, as well as a smaller difference by age \cite{lastofusarticle, lastofusimdb}. Such discrepancies may also be useful when understanding the different drivers of this phenomenon, and different user clusters can be profiled.

Finally, the rise of Large Language Models (LLMs) is changing the paradigm for better and worse. LLMs can be used to enhance and automate fake review detection \cite{adelani2020, G2021}. But on the other hand, LLMs are also becoming exceptionally good as quick, human-like content generators which can flood platforms such as these with review bombing and an elaborated text which can be challenging to label as machine-generated \cite{li2023, Stiff2022}. In this sense, LLMs are a new player which can be both part of the problem and the solution.

\backmatter



\bigskip
{\large \textbf{Data Availability}}
\newline
The dataset analyzed in this paper is publicly available at \href{https://www.kaggle.com/datasets/beridzeg45/metacritic-pc-games-reviews}{Kaggle}.

\bibliography{sn-bibliography}

\end{document}